\documentclass[runningheads]{llncs}
\pdfoutput=1
\usepackage[T1]{fontenc}
\usepackage[marginal]{footmisc}

\usepackage{cite}
\usepackage{amsmath,amssymb,amsfonts}
\usepackage{graphicx}
\usepackage{booktabs}
\usepackage{subcaption}
\usepackage{textcomp}
\usepackage{chemfig}
\usepackage[linesnumbered,ruled,vlined]{algorithm2e}
\usepackage{xcolor}
\usepackage{float}
\usepackage{hyperref}
\usepackage{circledsteps}
\definecolor{mygreen}{RGB}{120, 190, 100}
\definecolor{myred}{RGB}{175, 0 , 0}
\usepackage[misc]{ifsym}
\newcommand{\corr}{(\Letter)}
\usepackage{mwe}
\hypersetup{
    colorlinks=true,
    linkcolor=myred,  
    citecolor=mygreen,  
    urlcolor=blue    
}
\usepackage{caption} 
\usepackage{xcolor}
\newcommand\SimilarityG{{\scshape MSSM}}

\newcommand\Graph{$\mathcal{G_{M}}$}
\begin{document}
%
\title{Molecular Graph Representation Learning via Structural Similarity Information}
\titlerunning{Representation Learning via Structural Similarity Information}
%

%
%
\author{Chengyu Yao\inst{1,2}\thanks{These authors contributed equally to this work} \and
Hong Huang\inst{1,3*}\ \and
Hang Gao\inst{1,2}\and
Fengge Wu\inst{1,2} \corr \and
Haiming Chen\inst{3} \and
Junsuo Zhao\inst{1,2}
}
\authorrunning{C. Yao and H. Huang et al.}
\institute{University of Chinese Academy of Sciences, 100081 Beijing, China\\
\and
National Key Laboratory of Space Integrated Information System, Institute of
Software Chinese Academy of Sciences, 100081 Beijing, China \and Key Laboratory of System Software (Chinese Academy of Sciences) and State Key Laboratory of Computer Science, Institute of Software, Chinese Academy of Sciences, 100190 Beijing, China \\
\email{\{huanghong, chm\}@ios.ac.cn}\\
\email{\{yaochengyu2023, gaohang, fengge, junsuo\}@iscas.ac.cn}\\}
\maketitle              

\tocauthor{Chengyu Yao\inst{*}, Hong Huang\inst{*}, Hang Gao, Fengge Wu, Haiming Chen, Junsuo Zhao}
\toctitle{Molecular Graph Representation Learning via Structural Similarity Information}
 
\begin{abstract}
    Graph Neural Networks (GNNs) have been widely employed for feature representation learning in molecular graphs. Therefore, it is crucial to enhance the expressiveness of feature representation to ensure the effectiveness of GNNs. However, a significant portion of current research primarily focuses on the structural features within individual molecules, often overlooking the structural similarity between molecules, which is a crucial aspect encapsulating rich information on the relationship between molecular properties and structural characteristics. Thus, these approaches fail to capture the rich semantic information at the molecular structure level. To bridge this gap, we introduce the \textbf{Molecular Structural Similarity Motif GNN (MSSM-GNN)}, a novel molecular graph representation learning method that can capture structural similarity information among molecules from a global perspective. In particular, we propose a specially designed graph that leverages graph kernel algorithms to represent the similarity between molecules quantitatively. Subsequently, we employ GNNs to learn feature representations from molecular graphs, aiming to enhance the accuracy of property prediction by incorporating additional molecular representation information. Finally, through a series of experiments conducted on both small-scale and large-scale molecular datasets, we demonstrate that our model consistently outperforms eleven state-of-the-art baselines. The codes are available at https://github.com/yaoyao-yaoyao-cell/MSSM-GNN.
\end{abstract}
\section{Introduction}
Molecular Representation Learning, a critical discipline in bioinformatics and computational chemistry, has witnessed significant advancements in recent years\cite{duvenaud2015convolutional,kolouri2020wasserstein,sun2020graph}. Accurate prediction of molecular properties and activities is essential for drug discovery\cite{huang2020caster}, toxicity assessment\cite{zhang2023artificial}, and other biochemical applications\cite{shen2019molecular}. Nowadays, molecular representation learning has been widely integrated with Graph Neural Networks (GNNs), which are powerful tools for processing graph data and have been successfully applied in the molecular domain\cite{bouritsas2022improving,yu2022molecular,hevapathige2023uplifting}. However, most existing GNNs use the basic molecular graphs topology to obtain structural information through neighborhood feature aggregation and pooling methods\cite{kipf2016semi,gao2019graph,liu2022spherical}. This leads them to overlook the comprehensive chemical semantics. 

To address this challenge, several emerging approaches have been proposed around molecular graphs.
Specifically, some approaches\cite{yu2022molecular, wu2023molformer} take only the atom-level or motif-level information in heterogeneous molecular graphs as GNNs’ input to recognize common subgraphs with special meanings. By identifying the significance of ring compounds in molecular structures, Zhu et al. \cite{zhu2023mathcal} propose the Ring-Enhanced Graph Neural Network ($\mathcal {O}$-GNN). 
Alternatively, other methods \cite{yang2022learning,inae2023motif,geng2023novo} represent the molecular using motif-aware models that consider properties of domain-specific motifs.
Furthermore, there exists a multitude of techniques\cite{maziarz2021learning,yu2023atom,atsango20223d} that center their focus on studying the relations among substructures to recognize critical patterns hidden in motifs and improve the reliability of molecular property prediction. 

\begin{figure}[htbp]
\centering
\includegraphics[width=1\textwidth]{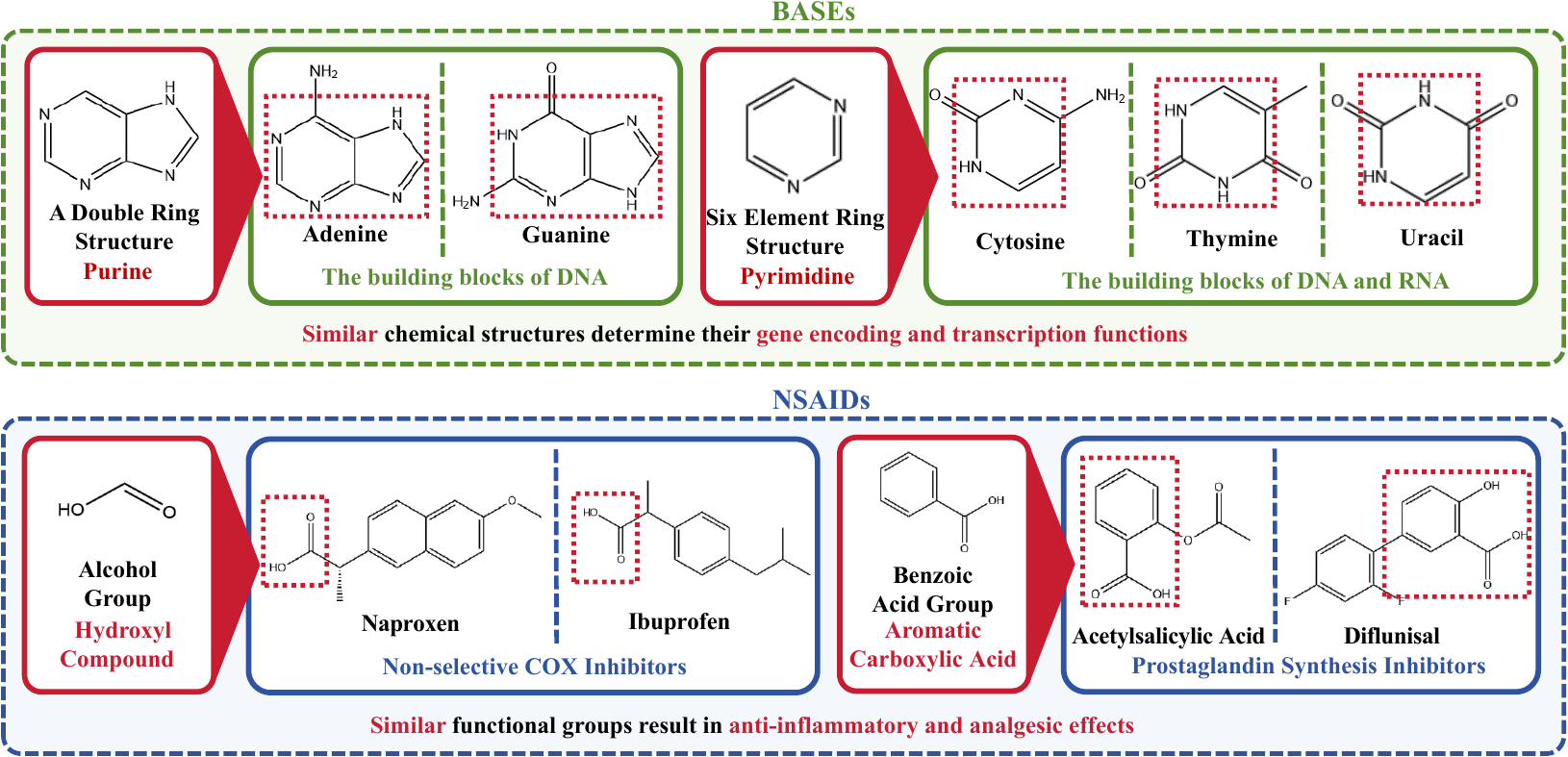}
\caption{Examples of molecules with similar structures often exhibit similar properties, a phenomenon observed in biological and chemical domains.}
\label{fig}
\end{figure}

Despite the considerable progress compared to traditional GNNs, most recent studies focus only on the message passing within individual molecules. The relationships between molecular structures are often ignored, which may result in the partial loss of semantic information. Moreover, the functions and properties of chemical molecules largely depend on their structures\cite{xue2000molecular}. For instance, consider examples illustrated in Fig.~\ref{fig}. Molecules with similar structures often have similar properties. Therefore, we need to take specific measures to represent the structural similarity between molecules, which can benefit downstream tasks such as molecular property prediction.

Based on the above-mentioned considerations, we design a \textbf{M}olecular \textbf{S}tructu \\ ral \textbf{S}imilarity \textbf{M}otif (MSSM) graph that empowers GNNs to capture the rich structural and semantic information from inter-molecule. The design starts by constructing a nested motif dictionary to re-represent molecular graphs. In light of the diverse node types present in motif-based molecular graphs, we propose a \textbf{M}ahalanobis \textbf{W}eisfeiler-\textbf{L}ehman \textbf{S}hortest-\textbf{P}ath (MWLSP) graph kernel. This kernel is designed to assess structural similarity from both the perspectives of length and position. It overcomes the limitation of the shortest path graph kernel\cite{borgwardt2005shortest}, which only retains connectivity information. By leveraging label information from different nodes and their neighbors, it provides a more granular representation of the graph, enhancing its expressiveness.

In this work, we propose a method that effectively considers inter-molecule structural similarity from a global perspective without sacrificing information in individual molecules. The method consists of the following major components: Firstly, it extracts motifs from molecules to create the motif dictionary and represents each molecule by utilizing the dictionary. Secondly, it uses our proposed Molecular Structural Similarity Motif (MSSM) graph to exploit rich semantic information from graph motifs. Finally, it applies GNNs to learn compositional and structural feature representation for each molecular graph and their similarities based on the MSSM graph. The experimental results show that our model can significantly outperform other state-of-the-art GNN models on various molecular property prediction datasets.

To summarize, our contributions are as follows:
\begin{itemize}
\item Considering the actual molecule structure, we designed a novel molecular graph representation method to represent motif structural information.

\item To improve the accuracy of GNNs in molecular property prediction tasks, we design a \SimilarityG\ graph by employing the MWLSP graph kernel. It quantifies the similarity between molecules through graph kernel scores and obtains a more comprehensive semantic representation at the structural level.
 
\item We show in the experiments that our model empirically outperforms state-of-the-art baselines on several benchmarks of real-world molecule datasets.

\end{itemize}

\section{Related Work}
\subsection{Motifs in Molecular Graphs}
Motif refers to the basic structure that constitutes any characteristic sequence. It can be viewed as a subgraph with a specific meaning in the molecular graph. For example, the edges in a molecular graph represent chemical bonds, and the rings represent the molecular ring structure. Several algorithms have been introduced to leverage motifs in different applications, including contrastive learning\cite{subramonian2021motif}, self-supervised pretraining\cite{zhang2021motif}, generation\cite{jin2020hierarchical} and drug-drug interaction prediction\cite{huang2020caster}. The motif extraction techniques used in the above methods, whether relying on exact counting\cite{cantoni2011morphological} or sampling and statistical estimation\cite{wernicke2006efficient}, have not utilized the structural similarities among motifs to enhance the expressiveness of molecular graphs.

\subsection{Molecular Graph Representation Learning}
DL has been widely applied to predict molecular properties. Molecules are usually represented as 1D sequences, including amino acid sequences, SMILES\cite{xu2017seq2seq} and 2D graphs\cite{duvenaud2015convolutional}. Wu et al.\cite{wu2022molecular} proposed a new molecular joint representation learning framework, MMSG, based on multi-modal molecular information (from SMILES and graphs). However, these approaches cannot capture the rich information in subgraphs or graph motifs. A few works based on GNNs have been reported to leverage motif-level information. Specifically, some approaches\cite{yu2022molecular,wu2023molformer,zhu2023mathcal} introduced the molecular graph representation learning method by constructing heterogeneous motif graphs from 
extracting different types of motifs. Alternatively, other methods\cite{maziarz2022learning,zhu2023mathcal} decomposed each training molecule into fragments by breaking bonds and rings in compounds to design novel GNN variants. Although these methods obtain more expressive molecular graphs, the challenge in motif-based approaches mainly comes from the difficulty in efficiently measuring similarities between input graphs. While existing graph kernel methods\cite{borgwardt2005shortest,shervashidze2011weisfeiler,hamilton2017inductive,dan2023self} can calculate scores by comparing different substructures of graphs to complete the measurement, there is currently no comparison method for motif-based molecular graphs. Therefore, our method focuses on learning motif structural information in the representation. 

\section{Methods}
\begin{figure*}[t]
\centering
\includegraphics[width=1\textwidth]{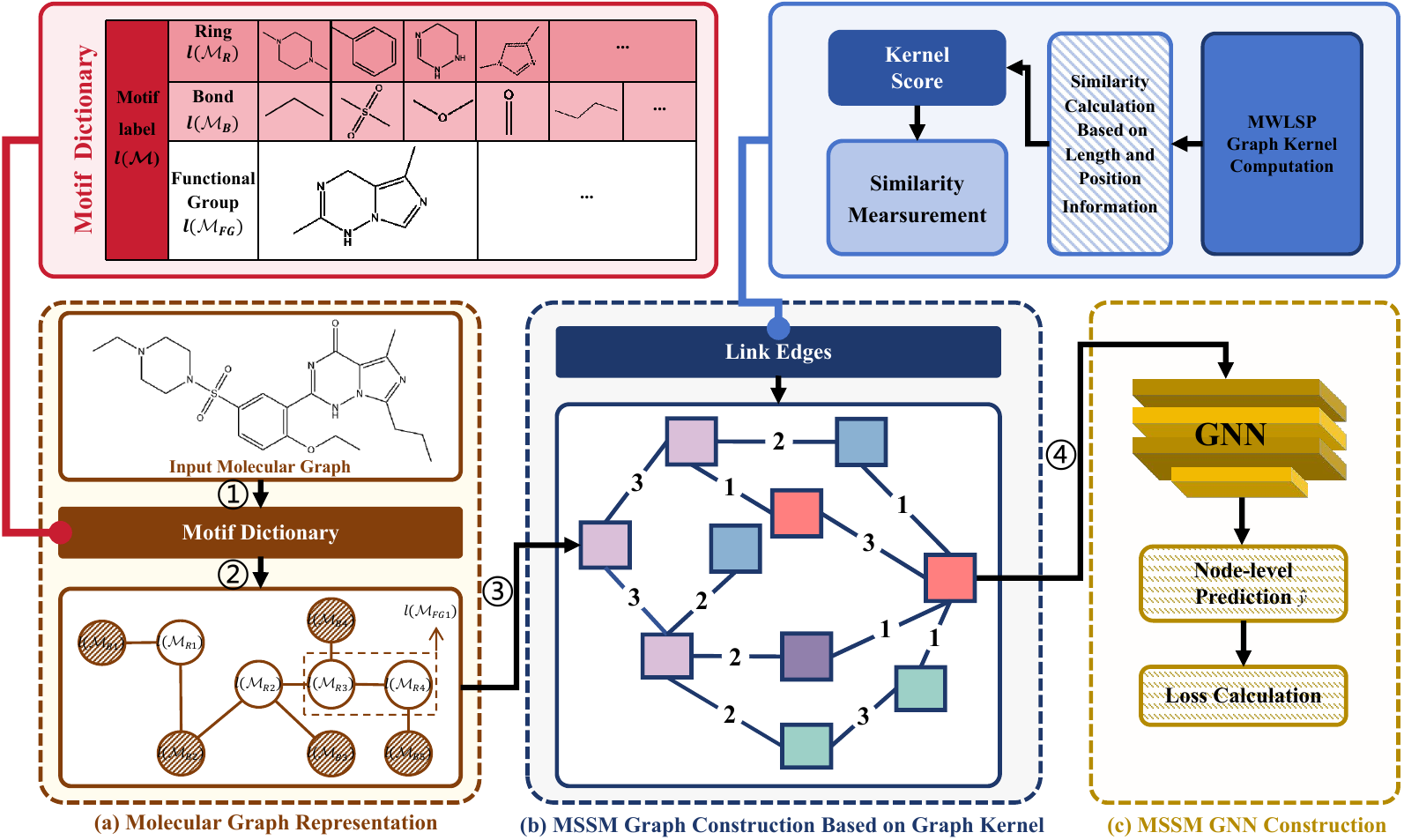} 
\caption{The framework of our proposed Molecular Structural Similarity Motif Graph Neural Network.}
\label{fig:framework}
\end{figure*}

In this section, we propose a novel method to construct a 
Molecular Structural Similarity Motif Graph Neural Network (MSSM-GNN)
(Illustrated in Fig.~\ref{fig:framework}) which takes the MSSM graph as input.

Generally, the framework of the method consists of three parts: $\textbf{(\romannumeral1)}$ Molecular graph representation; $\textbf{(\romannumeral2)}$ MSSM graph construction based on graph kernel; $\textbf{(\romannumeral3)}$ MSSM-GNN construction. Below, we explain in more detail about these parts. 
\subsection{Molecular Graph Representation}
In molecular graphs, motifs are subgraphs that appear repeatedly and are statistically significant. Therefore, we propose a novel molecular graph representation method based on chemical domain knowledge and  BRICS\footnote{breaking retrosynthetically interesting chemical substructures\cite{degen2008art}} algorithm to represent molecular structural information better. It considers both the internal atomic structure and the overall impact of special functional groups in the molecule. Its main process consists of the following two steps: \textbf{(\romannumeral1)} Motif Dictionary; \textbf{(\romannumeral2)} Molecular Graph Re-representation. 

\subsubsection{Motif Dictionary}
Let $\mathcal{G}=(V, E)$ denote a molecular graph, where $V$ is a set of atoms, $E \subseteq V \times V$ is a set of bonds between atoms. Generally, we denote a motif of the molecule $\mathcal{G}$ by $\mathcal{M} =<\hat{{V}},\hat{{E}}>$ , where $\hat{{V}}$ is a subset of ${V}$ and $\hat{{E}}$ is the subset of ${E}$ corresponding to $\hat{{V}}$, which includes all edges connecting nodes in $\hat{{V}}$. Due to the impact of ring, bond, and functional group structures on a molecule's stability, mechanical properties, and reactivity\cite{xue2000molecular}, we extract these structures as three distinct types of motifs from $\mathcal{G}$. This extraction aims to establish a correlation between molecular structure and properties, facilitating a targeted capture of diverse chemical features within the molecule. It considers important structural components within the molecule as much as possible and can be extended to different types of molecules, making it a general approach.

To systematically organize and store the extracted motif information, we construct a motif dictionary $\mathcal{D}$ by preprocessing all molecules in the dataset, as outlined in step $\textcircled{1}$ of Fig.~\ref{fig:framework}(a).
The $\mathcal{D}$ contains molecular identifiers as outer keys, each associated with nested dictionaries. These inner dictionaries categorize structural motif types with corresponding lists of extracted labels. We define the label $l(\mathcal{M})$ as the type of $\mathcal{M}$. The example in Fig.~\ref{fig:framework}(a) illustrates that ring type Piperazine\footnote{a six-membered ring compound containing two nitrogen atoms} can be expressed as $l(\mathcal{M}_{R1}$). This organization efficiently stores and retrieves structural information within each molecule.

\subsubsection{Molecular Graph Re-representation}
Based on the motif dictionary, we traverse the structure type and their corresponding motif lists for each molecule within it. This process aims to re-represent the molecular graph by establishing connections between the motifs in molecules. We defined the graph as $\mathcal{G}_\mathcal{M}=(\mathcal{V},\mathcal{E})$, where $\mathcal{V}$ and $\mathcal{E}$ denote a set of motifs and a set of adjacency relationship between motifs of the molecule, respectively. In the \Graph, a motif $\mathcal{M}$ is associated with a label $l(\mathcal{M})$ and adjacent motifs are connected by edges. As illustrated in step $\textcircled{2}$, for the drug molecule vardenafil, we can use the proposed algorithm to construct a $\mathcal{G}_\mathcal{M}$ from motifs out of the $\mathcal{D}$.

\subsection{MSSM Graph Construction Based on Graph Kernel}
Through the above method, we obtain a re-representation molecular graph \Graph. To provide GNN with more information, we will construct a Molecular Structural Similarity Motif (MSSM) graph in step $\textcircled{3}$. In the MSSM graph, each node represents a  $\mathcal{G}_\mathcal{M}$, and the edge represents two nodes $\mathcal{G}_{\mathcal{M}_1}$ and $\mathcal{G}_{\mathcal{M}_2}$ with structural similarity. We calculate the similarity between two $\mathcal{G}_\mathcal{M}$ by utilizing \textbf{M}ahalanobis \textbf{W}eisfeiler-\textbf{L}ehman \textbf{S}hortest-\textbf{P}ath(MWLSP) graph kernel. 

The fundamental idea of the graph kernel is to measure the similarity via the comparison of $\mathcal{G}_\mathcal{M}$' substructures. The kernel we proposed retains expressivity and is still computable in polynomial time. 

As depicted in Fig.~\ref{fig:framework}, MWLSP graph kernel takes $\mathcal{G}_{\mathcal{M}_1}$, $\mathcal{G}_{\mathcal{M}_2}$ as input, and its main process consists of the following steps: \textbf{(\romannumeral1)} Preprocess Input; \textbf{(\romannumeral2)} Perform MWLSP Graph Kernel Computation; \textbf{(\romannumeral3)} Comparison Scores of Graph-substructures. We give a pseudocode description of the MWLSP Graph Kernel in Algorithm~\ref{alg:myalgorithm}.

\subsubsection{Preprocess Input} 
In line~\ref{line:Ft}, we utilize Floyd-transformation (For detailed explanation, see \textbf{Appendix} A.1) $Ft(\mathcal{G}_\mathcal{M})$ to convert graphs $\mathcal{G}_{\mathcal{M}_1}$ and $\mathcal{G}_{\mathcal{M}_2}$ into graphs $\mathcal{G}_{F1}$ and $\mathcal{G}_{F2}$, respectively. $Ft(\mathcal{G}_\mathcal{M})$ generates the shortest path between all nodes in $\mathcal{G}_\mathcal{M}$. The shortest path between the vertex $v$ and $u$ is represented as $(v,u)$. The $(v,u)$ is the shortest path among all paths between two nodes. $\mathcal{G}_{F1}$ and $\mathcal{G}_{F2}$ contain all the information regarding the shortest path substructure partitions in $\mathcal{G}_{\mathcal{M}_1}$ and $\mathcal{G}_{\mathcal{M}_2}$, respectively. Specifically, $\mathcal{G}_{F1}$ has the same vertices as $\mathcal{G}_{\mathcal{M}_1}$, and the edge $(v, u)$ in $\mathcal{G}_{F1}$ represents detailed information about the shortest path in $\mathcal{G}_{\mathcal{M}_1}$. 

\begin{equation} \label{eq:Ft}
\mathcal{G}_{F1} = Ft(\mathcal{G}_{\mathcal{M}_1}) \qquad\qquad   \mathcal{G}_{F2} = Ft(\mathcal{G}_{\mathcal{M}_2})
\end{equation}

\subsubsection{Perform MWLSP Graph Kernel Computation} 
$K_{mwlsp}(\mathcal{G}_{F1}, \mathcal{G}_{F2})$ will compute the similarity between two graphs, $\mathcal{G}_{\mathcal{M}_1}$ and $\mathcal{G}_{\mathcal{M}_2}$, by summing up $k(e_1, e_2)$ i.e., the comparison scores between substructures $e_1$ and $e_2$ 
 in line~\ref{line:mwlsp}\textcolor{myred}{\texttt{-}} \ref{line:kernel}. $E_1'$ is the set of all edges in $\mathcal{G}_{F1}$ and $e_1$ is one of the edges in $E_1'$. $e_1$ represents a shortest path substructure in $\mathcal{G}_{\mathcal{M}_1}$, and similarly for $e_2$. 

\begin{equation} \label{eq:k_mwlsp}
K_{mwlsp}(\mathcal{G}_{F1}, \mathcal{G}_{F2}) = \sum_{e_1 \in E_1'} \sum_{e_2 \in E_2'} k(e_1, e_2)
\end{equation}

\begin{center}
\begin{algorithm}[H]
\small
\caption{MWLSP Graph Kernel Calculation}\label{alg:myalgorithm}
\KwData{$Graphs$ $\mathcal{G}_{\mathcal{M}_1}=(\mathcal{V}_{1},\mathcal{E}_{1})$, $\mathcal{G}_{\mathcal{M}_2}=(\mathcal{V}_{2}, \mathcal{E}_{2})$, $c$, $H$}
\KwResult{$Kernel \ Score$ $K_{mwlsp}$}

\SetKwFunction{MWLSPGraphKernel}{MWLSPGraphKernel}
\SetKwFunction{k}{k}
\SetKwFunction{LengthSim}{LengthSim}
\SetKwFunction{PositionSim}{PositionSim}
\SetKwProg{Fn}{Function}{:}{}

\Fn{\MWLSPGraphKernel{$\mathcal{G}_{\mathcal{M}_1}$, $\mathcal{G}_{\mathcal{M}_2}$, $c$, $H$}}{
    $\mathcal{G}_{F1} \gets Ft(\mathcal{G}_{\mathcal{M}_1})$ 
    $\mathcal{G}_{F2} \gets Ft(\mathcal{G}_{\mathcal{M}_2})$\;\label{line:Ft}
    
    $kernel\_score \gets 0$\;\label{line:mwlsp}

    \For{$e_1$ in $E(\mathcal{G}_{F1})$}{
        \For{$e_2$ in $E(\mathcal{G}_{F2})$}{
            $kernel\_score \mathrel{+}= k(e_1, e_2, c, H)$\;
        }}\KwRet $kernel\_score$\;\label{line:kernel}
}

\Fn{\LengthSim{$e_1$, $e_2$, $c$}}{\label{line:lengthsim}
    $sim_1 \gets\ $$max$$(0, c - \left| length(e_1) - length(e_2) \right|)$\;
    
    \KwRet $sim_1$\;\label{line:sim1}
}

\Fn{\PositionSim{$e_1$, $e_2$, $H$}}{\label{line:positionsim}
    $Initialize \ labels \ L_1 \ and \ L_2 \ based \ on \ e_1 \ and \ e_2$\;\label{line:initialize}
    
    \For{$h$ in $[0, H]$}{\label{line:propagate}
        \For{$u$ in $V(e_1)$}{\label{line:md}
            $nbrs\_sorted \gets sort(labels \ of \ neighbors \ of \  u \ lexicographically)$ \\
            $L^{(h+1)}(u) \gets hash(L^h(u), nbrs\_sorted)$
        }\label{line:hash}
        \For{$v$ in $V(e_2)$}{\label{line:md}
            $Calculate \ L^{(h+1)}(v) \ using \ the \ same \ method \ as \ above.$
        }\label{line:hash2}
            $Calculate \ the \ Mahalanobis \ distance \ D^{(h)}(u,v) \ between $ \\  $L^{(h)}(u) \ and \ L^{(h)}(v) \ at \ the \ h\texttt{-}th \ iteration.$
        }\label{line:md}
        $Sum \ D(u,v) \ across \ all \ final \ iteration \ yields \ sim_2$\label{line:sum}
    
    \KwRet $sim_2$\;\label{line:sim2}
}
\Fn{\k{$e_1$, $e_2$, $c$, $H$}}{\label{line:k}
    $sim_1 \gets \textbf{LengthSim}(e_1, e_2, c)$ $sim_2 \gets \textbf{PositionSim}(e_1, e_2, H)$\;
    
    \KwRet $sim_1 \times sim_2$\;\label{line:sim12}
}
\end{algorithm}
\end{center}
\begin{proposition}
\textit{Let n be the average number of nodes and d be the dimensionality of the features. Each node is associated with a $d$-dimensional feature vector. The time complexity for the kernel given by Eq.~\ref{eq:k_mwlsp} is $O(n^3 + n^4*(1+Hnd^3))$}.
\end{proposition}
The proof is given in the \textbf{Appendix} B.
\subsubsection{Comparison Scores of Graph-substructures}\label{method 3} For $k(e_1, e_2)$, we will calculate the similarity of substructures from two aspects: length and position. The calculation formulas are respectively $sim_{1}(e_1, e_2)$ and $sim_{2}(e_1,e_2)$.
For the aspect of length, $sim_{1}(e_1, e_2)$ utilizes the Brownian bridge\cite{chow2009brownian} to assess the similarity between $e_1$ and $e_2$ in line~\ref{line:lengthsim}\textcolor{myred}{\texttt{-}}\ref{line:sim1}. It returns the largest value when two edges have identical lengths and 0 when the edges differ in length by more than a hyperparameter \(c\). Furthermore, we can change the $c$ to control the similarity threshold, thus adjusting the filtering criteria. 

\begin{equation}
sim_{1}(e_1,e_2)=max(0, c - |length(e_1) - length(e_2)|)
\end{equation}

For the aspect of positional information, $sim_{2}(e_1,e_2)$ establishes a Weisfeiler-Lehman(WL) propagation scheme \cite{shervashidze2011weisfeiler} on the graphs, iteratively comparing labels on the nodes and their neighbors via Mahalanobis Distance(MD)\cite{de2000mahalanobis}. 

Specifically, we let $h$ be the current WL iteration which ranges from 0 to $H$($H$ is the total number of iterations). $L^h(u)$ is a set of node labels, representing the positional information of node $u$ at the current iteration $h$. $\mathcal{N}^h(u) = \{L^{\,h}(u_{left}), L^{\,h}(u_{right})\}$ represents the positional information of $u$'s neighboring nodes at the current iteration $h$. In the shortest path graph, $u_{left}$ and $u_{right}$ are the only two neighbor nodes of $u$. The scheme primarily consists of several steps, described in line~\ref{line:positionsim}\textcolor{myred}{\texttt{-}}\ref{line:sim2}:

Firstly, we compare two paths, $e_1$ and $e_2$, by utilizing the motif labels to initialize the sets of all node labels on these paths in line~\ref{line:initialize}.

\begin{equation}
L^{\,0}(u)=l(u)
\end{equation}

Next, if identical node labels exist, further iterative evaluation is conducted. We define the iterative rule with the hash function: in each iteration, the positional information of $u$ includes one more iteration of node connectivity compared to the previous iteration. By inputting the positional information of the current iteration's node $u$, i.e., $L^{\,h}(u)$ and its neighboring nodes, i.e, $\mathcal{N}^{\,h}(u)$, we use Eq.~\ref{eq:L} to compute $L^{\,h+1}(u)$, i.e., the positional information of $u$ in the next iteration $h+1$. And $\text{sort}(\cdot)$ sorts the labels lexicographically. The specific execution process is shown in line~\ref{line:propagate}\textcolor{myred}{\texttt{-}}\ref{line:hash2}.

\begin{equation}\label{eq:L}
\begin{split}
L^{\,h+1}(u)=hash(L^{\,h}(u), sort((L^{\,h}(v_1),..., L^{\,h}(v_{|\mathcal{N}(u)|})))),\\
v_j \in \mathcal{N}^{\,h}(u)), j \in \{1,...,|\mathcal{N}(u)|\}.
\end{split}
\end{equation}

Through the aforementioned process, we can represent the positional information of all nodes in $e_1$ and $e_2$ by utilizing $L^{\,h}(u)$.
Furthermore, in line~\ref{line:md}, we use MD (Please see \textbf{Appendix} A.2 for explanation) to measure the similarity between nodes. $D^{\,h}(u, v)$ denotes the MD  between the $L^{\,h}(u)$ and $L^{\,h}(v)$ at a specific iteration $h$. $M^{\,h}$ is the covariance matrix $Cov(L^{\,h}(u),L^{\,h}(v))$. The utilization of MD considers the diverse distribution characteristics of nodes belonging to different types in the heterogeneous feature space. In the context of $\mathcal{G}_\mathcal{M}$, distinct types of motifs may correspond to varied structures or properties. Therefore, we can quantify the similarity between motifs based on the distribution characteristics of each motif type. 

\begin{equation}
D^{\,h}(u, v) = \sqrt{(L^{\,h}(u) - L^{\,h}(v))^T M^{\,h} (L^{\,h}(u) - L^{\,h}(v))} 
\end{equation}

Finally, we cumulatively aggregate the MD from the 0-th to the $H$-th iteration in line~\ref{line:sum}. Through a weighted synthesis, we calculate the relational similarity between $u$ and $v$, considering positional information across all iterations. Therefore, we can calculate the similarity score $sim_{2}(e_1,e_2)$ by comparing the position similarity relationships among all nodes in $e_1$ and $e_2$. 

\begin{equation}
sim_{2}(e_1,e_2) = \sum_{u \in V(e_1)} \sum_{v \in V(e_2)} \exp\left(-\frac{1}{2} \sum_{h=0}^{H} D^{\,h}(u, v)\right) 
\end{equation}

For the above iterative process, we set two termination conditions:

\textbf{(\romannumeral1)} There is no intersection in the positional information of all nodes in $e_1$ and $e_2$ within the current iteration. This condition implies that, in the next iteration, the positional information of nodes for $e_1$ and $e_2$ is dissimilar, so we can terminate early.

\textbf{(\romannumeral2)} We have calculated the positional information for all iterations in $e_1$ and $e_2$, and the total number of iterations will not exceed $min(|\text{length}(e_1), \text{length}(e_2)|)$.

In summary, the comparison score of the graph substructure can be obtained by multiplying the similarities of the above two parts in line~\ref{line:k}\textcolor{myred}{\texttt{-}}\ref{line:sim12}.
\begin{equation}\label{method k}
k(e_1, e_2)= sim_{1}(e_1,e_2)*sim_{2}(e_1,e_2)
\end{equation}

\subsubsection{MSSM Graph Construction} 
We can construct the MSSM graph based on the similarity calculation result of the above MWLSP graph kernel. Since the structural similarity analysis of molecules often does not require very precise numerical values, it focuses on the relative similarity between molecules. To reduce the complexity of the comparison, we simplify the kernel score to an integer range of [0, 3] by dividing it by the maximum achievable value:

\begin{equation}
S(Ft(\mathcal{G}_{Mi}), Ft(\mathcal{G}_{Mj})) = \left\lfloor \frac{3 \cdot K_{mwlsp}(Ft(\mathcal{G}_{Mi}), Ft(\mathcal{G}_{Mj}))}{max(K_{mwlsp}(Ft(\mathcal{G}_{Mi}), Ft(\mathcal{G}_{Mj})))} \right\rfloor
\end{equation}

where $\left\lfloor{x}\right\rfloor$ represents rounding $x$ down to the nearest integer.

Considering the above possible calculation results, we use the similarity score $S(Ft(\mathcal{G}_{Mi}), Ft(\mathcal{G}_{Mj}))$ to represent the corresponding edge weight value $A_{ij}$ and formally establish detailed measurement standard $\text{Sim}_{ij}$ as follows:

\begin{equation}
Sim_{ij} = \left\{
\begin{array}{ll}
    Very \ High \ Similarity & {       if       A_{ij}=3, }  \\
    Relatively \ High \ Similarity & {       if       A_{ij}=2, }  \\ 
    Average \ Similarity \ & {       if       A_{ij}=1, }  \\
    Dissimilar & {       if       A_{ij}=0}
\end{array}
\right.
\end{equation}

where if $A_{ij} > 0$, $\mathcal{G}_{Mi}$ and $\mathcal{G}_{Mj}$ have a similar relationship, and a connecting edge with corresponding weight value needs to be established; otherwise, there is no need to perform connection processing. Fig.~\ref{fig:framework}(b) provides an example of MSSM graph construction.

\subsection{MSSM-GNN Construction}
In this part, we build an MSSM-GNN to learn graph structural feature representations of the MSSM graph. In graph learning, the input MSSM graphs can be denoted as $\mathcal{G}_{MSSM}=(\mathcal{V}_{MSSM},\mathcal{E}_{MSSM})$, where $\mathcal{V}_{MSSM}$ is the node set of $\mathcal{G}_\mathcal{M}$, and $\mathcal{E}_\mathcal{MSSM}$ is the edge set of similarity relationship between two $\mathcal{G}_\mathcal{M}$. And we use $y \in \mathcal{Y}$ as the node-level property label for $\mathcal{G}_{\mathcal{M}_i}$, where $\mathcal{Y}$ represents the label space. 

For graph property prediction, a predictor with the encoder-decoder architecture is trained to encode $\mathcal{G}_{MSSM}$ into a node representation vector in the latent space and decode the representation to predict ${\hat{y}}$. Specifically, we fed the MSSM graph data into GNN to acquire ${\hat{y}}$ (corresponds to step $\textcircled{4}$):

\begin{equation}
\hat{y}=\text{GNN}(\mathcal{G}_{MSSM})\in \mathcal{Y}.
\end{equation}

The loss function used in our model is the label prediction loss. The label prediction loss function $\mathcal{L}_{pred}$ is derived similarly to existing methods:

\begin{equation}
\mathcal{L}_{pred}=\text{CE}(\hat{y},y).
\end{equation}

where $\hat{y}$ represents the predicted value, $y$ is the ground truth, and $\text{CE}$ represents the Cross-Entropy loss function used in classification tasks.

In this way, we can get a more comprehensive feature representation of the entire $\mathcal{G}_{MSSM}$. It contains all the information on the connected motifs, retaining the atomic structure relationships and connections within the original motifs. Therefore, we can get a more accurate prediction of molecular properties based on the MSSM-GNN. The process is illustrated in Fig.~\ref{fig:framework}(c).

\section{Experiments}
In this section, we investigate how our proposed method improves GNN performance on molecular property tasks. In our investigations, we raise the following questions: \textbf{Q1}: Compared with state-of-the-art baselines, how effective is MSSM-GNN in improving the accuracy of molecular prediction on common bioinformatics graph benchmark datasets? \textbf{Q2}: If experiments are conducted on real-world datasets, will MSSM-GNN still have an effect? \textbf{Q3}: Does feature learning of similarities between molecules play a more critical role in MSSM-GNN? \textbf{Q4}: What impact will the setting of the similarity threshold on different datasets have on the final classification results?

In response to the above problems, we conducted a series of experimental studies. Some basic settings of experiments and analysis of results are as follows:

\subsection{Experimental Settings}
\subsubsection{Datasets.}
To verify whether MSSM-GNN provides more information conducive to accurate classification, we evaluate our model on five popular bioinformatics graph benchmark datasets from TUDataset\cite{morris2020tudataset}, which includes four molecular datasets PTC\cite{toivonen2003statistical}, MUTAG\cite{debnath1991structure}, NCI1\cite{wale2008comparison}, MUTAGENICITY\cite{kazius2005derivation} and one protein dataset PROTEINS\cite{borgwardt2005protein}.
\begin{table*}[ht]\scriptsize
    \footnotesize  
	\setlength{\tabcolsep}{2pt}
 	\caption{Graph classification accuracy (\%) on various TUDataset graph classification tasks. The best performers on each dataset are shown in \textbf{bold}.}
	\label{tab:mole}
	\begin{center}
		\begin{tabular}{l|cccccc}
			\hline\rule{-2pt}{10pt}		
			\textbf{Methods} &  \textbf{PTC}  & \textbf{NCI1} & \textbf{MUTAG} & \textbf{PROTEINS} & \textbf{MUTAGENICITY}
            \\
			\hline\rule{-2pt}{10pt}
   			\text{DGCNN} & 58.6$\pm$2.5 & 74.4$\pm$0.5 & 85.8$\pm$1.7 & 75.5$\pm$0.9 & 72.3$\pm$2.6	 \\
			\text{GCN} & 64.2$\pm$4.3 & 80.2$\pm$2.0 & 85.6$\pm$5.8 & 76.0$\pm$3.2 & 79.8$\pm$1.6	 \\
			\text{GIN} & 64.6$\pm$7.0 & 82.7$\pm$1.7 & 89.4$\pm$5.6 & 76.2$\pm$2.8 & 82.0$\pm$0.3	 \\
                \text{PatchySAN} & 60.0$\pm$4.8 & 78.6$\pm$1.9 & 92.6$\pm$4.2 & 75.9$\pm$2.8 & 77.9$\pm$1.3	 \\
			\text{GraphSAGE} & 63.9$\pm$7.7 & 77.7$\pm$1.5 & 85.1$\pm$7.6 & 75.9$\pm$3.2 & 78.8$\pm$1.2	 \\
			\text{PPGN} & 66.2$\pm$6.5 & 83.2$\pm$1.1 & 90.6$\pm$8.7 & 77.2$\pm$4.7 & 78.6$\pm$0.9	 \\
 			\text{WEGL} & 64.6$\pm$7.4 & 76.8$\pm$1.7 & 88.3$\pm$5.1 & 76.1$\pm$3.3 & 80.8$\pm$0.4	 \\
			\text{CapsGNN} &  71.2$\pm$1.9 & 78.4$\pm$1.6 &  86.7$\pm$6.9& 76.3$\pm$4.6 & 79.5$\pm$0.7	 \\
    		\text{GSN} & 68.2$\pm$7.2 & 83.5$\pm$2.3& 90.6$\pm$7.5 & 76.6$\pm$5.0 & 81.0$\pm$1.5	 \\
			\text{HM-GNN} & 78.5$\pm$2.6 & 83.6$\pm$1.5 & 96.3$\pm$2.8 & 79.9$\pm$3.1 & 83.0$\pm$1.1	 \\
			\text{GPNN} & 78.2$\pm$1.2 & 83.1$\pm$0.3 & 92.6$\pm$1.8 & 76.8$\pm$3.9 & 83.0$\pm$0.4 \\   
            \hline\rule{-2pt}{10pt}
   			\textbf{OURS}   &  \bf{81.1$\pm$1.7} &  \bf{85.5$\pm$0.3} &   \bf{97.3$\pm$2.6}  & \bf{83.3$\pm$0.4} & \textbf{84.0$\pm$0.5}  \\ 
			\hline 
		\end{tabular}
    \vskip -0.1in
	\end{center}
\end{table*}
\subsubsection{Baselines.}
We compare our model with eleven state-of-the-art GNN models for molecular property tasks: Deep Graph CNN (DGCNN)\cite{phan2018dgcnn}, GCN\cite{kipf2016semi}, GIN \cite{xu2018powerful}, 
PATCHYSAN \cite{niepert2016learning}, GraphSAGE\cite{hamilton2017inductive}, Provably Powerful Graph Networks (PPGN)\cite{maron2019provably}, Wasserstein Embedding for Graph Learning (WEGL)\cite{kolouri2020wasserstein}, Capsule Graph Neural Network (CapsGNN)\cite{xinyi2018capsule}, GSN\cite{bouritsas2022improving}, HM-GNN\cite{yu2022molecular}, GPNN\cite{hevapathige2023uplifting}.

\subsection{Performance Evaluation on Molecular Graph Datasets}
To learn graph feature representations in our molecular structural similarity motif graphs, 3 GNN layers are applied. For a fair comparison, we evaluate all baselines using the experiment settings provided by \cite{yu2022molecular}. The hyper-parameters we tune for each dataset are (1) the learning rate$ \in {0.01, 0.05}$; (2) the number of hidden units$ \in {16, 64, 1024}$; (3) the dropout ratio$ \in {0.2, 0.5}$. We set the verification method as the mean and standard deviation of the seven best validation accuracies from ten folds. We compare MSSM-GNN with the baseline approaches on the abovementioned dataset to answer \textbf{Q1}. 
The comparison results are summarized in Table~\ref{tab:mole}. We make the following observations:

\textit{MSSM-GNN significantly outperforms baseline models on all five datasets for molecular prediction.} Among them, on the PROTEINS dataset, the accuracy of MSSM-GNN increased by 3.4\% compared with the best method. The superior performances on five molecular datasets demonstrate that motif substructures extracted from the motif dictionary, along with the calculated similarity relationships between molecular nodes based on it, facilitate GNNs in learning improved motif-level and molecular-level feature representations of molecular graphs.
\begin{table}[ht]
    \scriptsize
    \footnotesize  
	\setlength{\tabcolsep}{3pt}
    \caption{Graph Classification Results (\%) on Open Graph Benchmark datasets. }
    \label{tab:large}
    \begin{center}
        \begin{tabular}{l|cccc} 
            \hline\rule{-3pt}{10pt}            
            \textbf{Methods} &  \textbf{ogbg-molhiv}  & \textbf{ogbn-proteins} & \textbf{ogbg-moltoxcast} & \textbf{ogbg-molpcba} \\
            \hline\rule{-3pt}{10pt}
            GCN & 75.99$\pm$1.19  & 72.51$\pm$0.35  & 61.13$\pm$0.47 & 24.24$\pm$0.34 \\
            GIN & 77.07$\pm$1.49  & 77.68$\pm$0.20  & 62.19$\pm$0.36 & 27.03$\pm$0.23 \\
            GSN & 77.90$\pm$0.10  & 85.80$\pm$0.28  & 62.61$\pm$0.45 & 27.00$\pm$0.70 \\
            PNA & 79.05$\pm$1.32  & 86.82$\pm$0.18  & 63.47$\pm$0.67 & 25.70$\pm$0.60 \\
            HM-GNN & 79.03$\pm$0.92 & 86.42$\pm$0.08 & 64.38$\pm$0.39 & 28.70$\pm$0.26 \\
            GPNN & 77.70$\pm$2.30   & 87.74$\pm$0.13  & 65.22$\pm$0.47 & 28.90$\pm$0.91 \\  
            \hline\rule{-3pt}{10pt}
            \textbf{OURS}   &   \bf{79.70$\pm$0.03}  &  \bf{89.17$\pm$0.07} &  \bf{66.57$\pm$1.00} &  \bf{30.07$\pm$0.37}\\
            \hline 
        \end{tabular}
        \vskip -0.1in
    \end{center}
\end{table}
\subsection{Performance Evaluation on Large-Scale Real-World Datasets}
To answer \textbf{Q2}, we evaluate our model on four large-scale real-world datasets from the Open Graph Benchmark (OGB)\cite{hu2020open}. They are two binary classification datasets-- ogbg-molhiv, ogbn-proteins and two multiclass classification datasets--ogbg-molt
oxcast, ogbg-molpcba. 

In this part, we compare our model with GIN, GCN, GSN, PNA, HM-GNN and GPNN. Except that the hyperparameters we tuned for each dataset varied as (1) learning rate$ \in {0.01, 0.001}$; (2) number of hidden units$ \in {10, 16}$; (3) dropout rate$ \in {0.5, 0.7, 0.9}$; 
(4) the batch size$ \in {128, 5000, 28000}$, others are the same as above experiment. Table~\ref{tab:large} shows the AP results on Ogbg-molpcba and ROC-AUC results on the other three datasets. We observe: \textit{our method is significantly better than the other compared methods by obvious margins.} The results prove our model’s superior generalization ability on real-world datasets, which is crucial for its potential applications in various domains, including drug discovery, bioinformatics, and chemical safety assessment.

\subsection{Ablation Study}
To address \textbf{Q3}, we conducted ablation experiments on different components of MSSM-GNN, focusing on the motif-based molecular graph representation and the similarity calculation. The corresponding conclusions are as follows:

\subsubsection{Effect of Motif-Dictionary Representation}
As shown in Table~\ref{tab:ab}, comparing task performance before and after removing the motif dictionary module yields the following observations: Performance on three graph classification datasets benefits from the module, resulting in accuracy improvements ranging from 0.8\% to 2.8\%. These improvements could potentially be attributed to the module's effective learning of valuable information about the molecule's substructure. 

\begin{table}[ht]\scriptsize
    \footnotesize  
	\setlength{\tabcolsep}{7pt}
 	\caption{Ablation studies of the motif dictionary and measurement of length and position similarity.}
	\label{tab:ab}
	\begin{center}
		\begin{tabular}{l|cccccc}
			\hline\rule{-3pt}{10pt}			
\textbf{Datasets} & \textbf{PTC\_MR} & \textbf{PTC\_FR} & \textbf{MUTAG} & \textbf{PROTEINS} \\
			\hline\rule{-3pt}{10pt}
			\textbf{MSSM-GNN} &\bf{81.1$\pm$1.7} & \bf{80.9$\pm$1.5 }&  \bf{97.3$\pm$2.6} & \bf{83.3$\pm$0.4}	 \\
                \text{w/o motif dictionary} &78.3$\pm$1.1  &78.6$\pm$1.4 &96.5$\pm$2.7  &82.5$\pm$1.2	 \\
                \text{w/o length similarity} &77.9$\pm$1.5  &78.3$\pm$1.1  &94.6$\pm$0.2  &81.2$\pm$0.9	 \\
                \text{w/ edit distance} &77.1$\pm$2.9  &78.2$\pm$1.7  &93.1$\pm$0.6  &80.8$\pm$0.4	 \\
			\hline 
		\end{tabular}
    \vskip -0.1in
	\end{center}
\end{table}
\subsubsection{Effect of Length-Similarity Calculation}
As shown in Table~\ref{tab:ab}, we observe a significant drop in performance when the length-similarity calculation is not included, amounting to an absolute drop of 2.1\% - 3.2\%. 
These observations confirm that evaluating path structures from a length perspective indeed facilitates the learning of the global information and inherent connectivity relationships among motif-level substructures, thereby contributing to representing graph information more comprehensively.

\subsubsection{Effect of Position-Similarity Calculation}
In MSSM-GNN, the location similarity calculation method we designed is MWL. By replacing MWL with edit distance, we examined the impact of the graph similarity metric. As Table~\ref{tab:ab} shows, MWL offers advantages over edit distance. MWL not only quantifies structural similarity but also incorporates the type and position information of different nodes in graph modeling, thereby effectively representing the real molecular graph structure. Meanwhile, MWL becomes particularly advantageous for larger-scale graph datasets, offering significant enhancements by extracting richer structural information. For example, our model enhances PROTEINS more than MUTAG.

\subsection{Sensitive Analysis}

\begin{figure}[t]
    \centering
    \begin{subfigure}{0.32\textwidth}
        \centering
        \includegraphics[width=\textwidth]{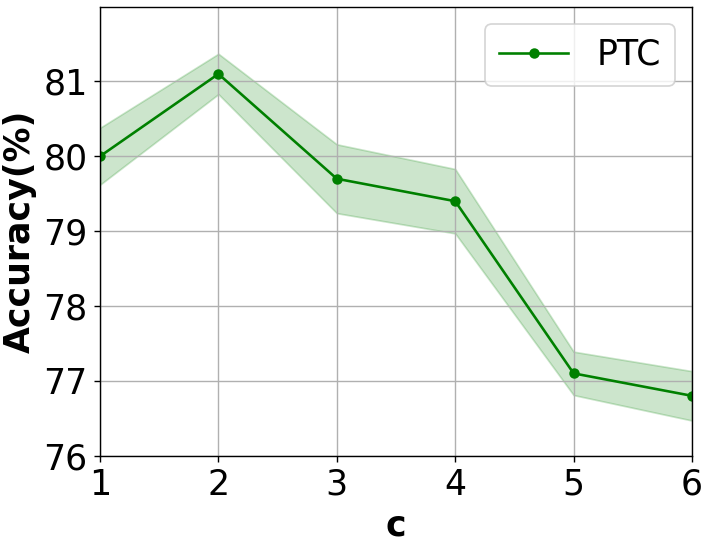}
        \caption{PTC}
        \label{fig:subfig_a}
    \end{subfigure}\hfill
    \begin{subfigure}{0.32\textwidth}
        \centering
        \includegraphics[width=\textwidth]{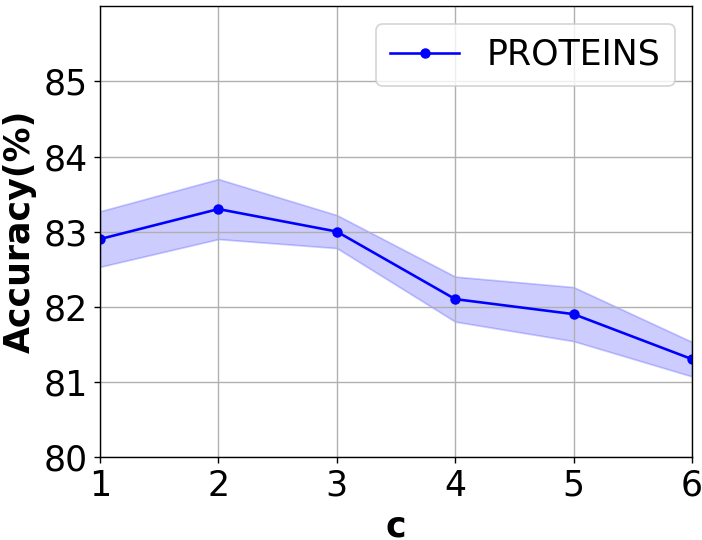}
        \caption{PROTEINS}
        \label{fig:subfig_b}
    \end{subfigure}\hfill
    \begin{subfigure}{0.32\textwidth}
        \centering
        \includegraphics[width=\textwidth]{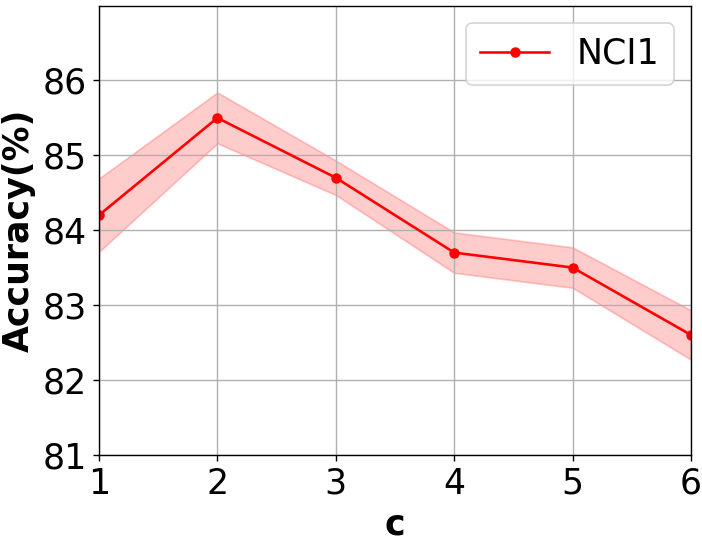}
        \caption{NCI1}
        \label{fig:subfig_c}
    \end{subfigure}
    \caption{Performance of MSSM-GNN on three different datasets with varying hyperparameters $c$.}
    \label{fig:charts}
\end{figure}

In this part, to explore \textbf{Q4}, we further evaluate the hyperparameter $c$ that we introduced in our proposed similarity calculation formula. We modify the value of hyperparameter $c$ that controls the similarity threshold and observe how the performance changes. We perform such experiments on multiple datasets. The results are shown in Fig.~\ref{fig:charts}.

As observed, the performance peaks when the value of $c$ is 2 across all three datasets. With the increase in $c$, the impact of the similarity threshold on training also becomes more pronounced. It is evident that the performance of MSSM-GNN decreases as $c$ increases from 2 to 6, indicating that the $c$ indeed influences the representation learning capabilities of MSSM-GNN. We believe that $c$ assists in filtering out samples with low similarity, emphasizing those contributing more significantly to the training, thereby enhancing overall performance.
\section{Conclusions}
This paper proposes an effective model for molecular graph representation learning, Molecular Structural Similarity Motif GNN (MSSM-GNN). We explicitly incorporate the similarity representations between molecules into GNN and jointly update them with motif representations. Specifically, we connect two molecules through edge weights calculated by a novel MWLSP graph kernel, enabling message passing between molecular graphs. We use the GNN model to learn the MSSM graph and get the motif-level and molecule-level graph embedding. Experiments demonstrate the superiority of our model in various datasets, which beats a group of baseline algorithms. 

\begin{credits}
\subsubsection{\ackname} 
The authors would like to thank the editors and reviewers for their valuable comments. This work is supported by the CAS Project for Young Scientists in Basic Research (Grant No. YSBR-040) and the National Natural Science Foundation of China (Grant No. 62372439),  the Natural Science Foundation of Beijing, China (Grant No. 4232038), the Basic Research Program of ISCAS (Grant No. ISCAS-JCMS-202307).

\end{credits}
\bibliographystyle{splncs04}

%

\appendix
\section{Explanations.} 
\subsection{Explanation of Floyd-transformation}
The Floyd transformation is a method for transforming a graph into its shortest-path graph. It is typically employed to solve shortest-path problems. Its concept is based on dynamic programming, gradually updating the shortest path information between nodes to obtain the shortest paths among all nodes in the graph. We give a pseudocode description of the Floyd transformation in Algorithm~\ref{alg:floyd}.

\subsection{Explanation of Mahalanobis distance}
The Mahalanobis distance is a metric used to measure the similarity or dissimilarity between two samples. It considers the correlations between individual features, thus providing a more accurate reflection of the actual distance between data points. Given two vectors or sample points, $\mathbf{x}$ and $\mathbf{y}$, their Mahalanobis distance can be defined as: 

\[ D_{M}(\mathbf{x}, \mathbf{y}) = \sqrt{(\mathbf{x} - \mathbf{y})^{\top} \cdot \Sigma^{-1} \cdot (\mathbf{x} - \mathbf{y})} \]

Where $(\mathbf{x} - \mathbf{y})^\top$ represents the transpose of the vector $(\mathbf{x} - \mathbf{y})$, $\Sigma^{-1}$ represents the inverse matrix of the covariance matrix $\Sigma$, the product denotes the matrix multiplication between the vector $(\mathbf{x} - \mathbf{y})$ and $\Sigma^{-1}$, and finally, taking the square root of the result gives the Mahalanobis distance.

\begin{algorithm}[H]
\caption{Floyd-transformation}\label{alg:floyd}
\KwData{$Graph$ $G=(V, E)$}
\KwResult{$Shortest Path Graph$ $G'=(V, E')$}

\SetKwFunction{FloydTransform}{FloydTransform}
\SetKwProg{Fn}{Function}{:}{}

\Fn{\FloydTransform{$G$}}{
    $n \gets |V|$ \\
    $D \gets graph.adjacencyMatrix$\\
    \For{$k \gets 1$ \KwTo $n$}{
        \For{$i \gets 1$ \KwTo $n$}{
            \For{$j \gets 1$ \KwTo $n$}{
                \If{$D[i][j] > D[i][k] + D[k][j]$}{
                    $D[i][j] \gets D[i][k] + D[k][j]$
                }
            }
        }
    }
    $E' \gets \emptyset$
    \For{$i \gets 1$ \KwTo $n$}{
        \For{$j \gets 1$ \KwTo $n$}{
            \If{$D[i][j] < \infty$}{
                $E' \gets E' \cup \{(i, j)\}$
            }
        }
    }
    \KwRet $G'=(V, E')$
}
\end{algorithm}
Suppose a graph $G = (V, E)$ consists of $n$ nodes, where $V$ represents the set of nodes, and $E$ represents the set of edges. Each node $v_i \in V$ has an associated feature vector $\mathbf{x}_i$, representing the node's label information. The Mahalanobis distance $D_M(v_i, v_j)$ between nodes $v_i$ and $v_j$ can be expressed as the Mahalanobis distance between the node label vectors $\mathbf{x}_i$ and $\mathbf{x}_j$.

\section{Proof of the complexity of MWLSP} 
This section provides the proof of Property 1 (Complexity of MWLSP).

\textit{Proof.} Let us assume that we are dealing with two graphs with $n$ nodes each. Each node is associated with a $d$-dimensional feature vector, where $d$ represents the dimensionality of the features.

In the first step, the Floyd transformation can be done in $O(n^3)$\cite{borgwardt2005shortest} when using the Floyd-Warshall algorithm. In the second step, we have to consider pairwise comparison of all edges in both transformed graphs. The number of edges in the transformed graph is $n^2$, resulting in a total runtime of $O(n^4)$.
In the third step, the calculation of length-based similarity is constant time, denoted by $O(1)$, owing to minimal mathematical operations. However, for positional information, the runtime complexity of the Weisfeiler-Lehman scheme with $H$ iterations is $O(Hn)$\cite{shervashidze2011weisfeiler}. Within each iteration, computing the Mahalanobis distance between nodes necessitates $O(d^3)$ time\cite{de2000mahalanobis}, resulting in a total time complexity of $O(Hnd^3)$.

In summary, considering each component's detailed complexity analysis, the algorithm's overall time complexity is as follows :$O (n^3 + n^4*(1+Hnd^3))$. Therefore, we can categorize the complexity of the entire algorithm as polynomial.

\end{document}